\pdfoutput=1
\documentclass[11pt]{article}

\usepackage{ACL2023}

\usepackage{times}
\usepackage{latexsym}
\usepackage{graphicx}
\usepackage{tabularx}
\usepackage{soul}
\usepackage{adjustbox}
\usepackage{hyperref}
\usepackage[T1]{fontenc}
\usepackage[utf8]{inputenc}

\usepackage{mathptm}

\usepackage{microtype}

\title{Improving Automatic Quotation Attribution in Literary Novels}

\author{Krishnapriya Vishnubhotla$^{1,4}$, Frank Rudzicz$^{2,4,1}$, Graeme Hirst$^{1}$ \and  Adam Hammond$^{3}$ \\
        \textsuperscript{1}Department of Computer Science, University of Toronto  \\ 
        \textsuperscript{2}Faculty of Computer Science, Dalhousie University \\ 
        \textsuperscript{3}Department of English, University of Toronto \\ \textsuperscript{4}Vector Institute for Artificial Intelligence \\ 
        \texttt{\{vkpriya,frank,gh\}@cs.toronto.edu adam.hammond@mail.utoronto.ca}}

\begin{document}
\maketitle
\begin{abstract}
    Current models for quotation attribution in literary novels assume varying levels of available information in their training and test data, which poses a challenge for in-the-wild inference. Here, we approach quotation attribution as a set of four interconnected sub-tasks: character identification, coreference resolution, quotation identification, and speaker attribution. We benchmark state-of-the-art models on each of these sub-tasks independently, using a large dataset of annotated coreferences and quotations in literary novels (the Project Dialogism Novel Corpus). We also train and evaluate models for the speaker attribution task in particular, showing that a simple sequential prediction model achieves accuracy scores on par with state-of-the-art models\footnote[1]{Code and data can be found at \url{https://github.com/Priya22/speaker-attribution-acl2023}}. 
\end{abstract}

\section{Introduction}
\label{sec-intro}
We focus on the task of automatic {\it quotation attribution}, or {\it speaker identification}, in full-length English-language literary novels. The task involves attributing each quotation (dialogue) in the novel to the character who utters it. The task is complicated by several factors: characters in a novel are referred to by various names and aliases ({\it Elizabeth, Liz, Miss Bennet, her sister}); these aliases can change and be added over the course of the novel;
% as character arcs and relationships progress; 
and authors often employ differing patterns of dialogue in the text, whereby quotations are sometimes attached to the speaker explicitly via a speech verb, and at other times require keeping track of character turns over multiple paragraphs. The development of automated methods has also been hindered by the paucity of annotated datasets on which models can be trained and evaluated. 

Existing methods for quotation attribution fall into one of two groups: those that directly attribute the quotation to a named character entity and those that treat it as a two-step process in which quotations are first attached to the nearest relevant \textit{mention} of a character and mentions are then resolved to a canonical character name via a coreference resolution model. We contend that most use-cases of a quotation attribution system involve resolving the speaker mention to one among a list of character entities. Thus, the usability of these systems is very much dependent on their ability to  compile such a list of character entities and to resolve each attributed mention to an entity from this list.`

Here, we use the Project Dialogism Novel Corpus \cite{vishnubhotla-etal-2022-project}, a large dataset of annotated coreferences and quotations in literary novels, to design and evaluate pipelines of quotation attribution. Our analysis shows that state-of-the-art models are still quite poor at character identification and coreference resolution in this domain, thus hindering functional quotation attribution.

\section{Background and Prior Work}
\label{sec-rel}
\newcite{elson2010automatic} introduce the CQSA corpus, which contains quotations from excerpts from 4 novels and 7 short-stories that are annotated for the nearest speaker mention, which can be named (e.g., \textit{Elizabeth}), or nominal (\textit{her friend}). On average, only 25\% of the attributions in CQSA are to a named entity. 

In contrast, \newcite{he2013identification} link quotations directly to entities, and a list of characters and aliases is required for attribution. This list is generated with a named entity recognition (NER) model to obtain entity terms, which are then grouped together using Web resources such as Wikipedia.

The GutenTag package from \newcite{brooke2015gutentag} contains modules for generating character lists and identifying speakers in literary texts. The former is based on the LitNER model \cite{brooke2016bootstrapped}, which bootstraps a classifier from a low-dimensional Brown clustering of named entities from Project Gutenberg texts. The speaker attribution model is a simple rule-based approach that identifies the nearest named entity. 

% The QuoteLi dataset from \newcite{muzny2017two} contains both quote--mention and mention--entity annotations. They evaluate the two modules separately, though again assuming that a list of named character entities is given.

\newcite{sims2020measuring} annotate the first 2000 tokens of 100 novels from the LitBank dataset\footnote{https://github.com/dbamman/litbank}. Quotations are linked to a unique speaker from a predefined list of entities. LitBank also contains annotations for coreference for these tokens \cite{bamman2020annotated}. The BookNLP package\footnote{https://github.com/booknlp/booknlp} from the same group contains pre-trained models for NER, coreference resolution, and speaker attribution, although the latter is only at the mention-level.

\newcite{cuesta2022does} attempt to reconcile the differences in pre-requisites and methodologies of prior attribution systems by proposing a modularization of the task into three sub-tasks: quotation identification, character identification, and speaker attribution. They evaluate baselines for each component, propose a new state-of-the-art method for speaker attribution, and quantify the relative importance of each module in an end-to-end pipeline. Their speaker attribution module, however, considers only named mentions in the text as candidate speakers, leading to a lower performance on implicit and anaphoric quotations. Neither their dataset of 15 novels nor their model for speaker attribution have been made public, precluding comparison with our work below.

In our work, we follow this modular formulation, with some key differences: (a) we evaluate an additional sub-task of coreference resolution, allowing us to (b) test an attribution model that can work with both named and pronominal candidate mentions surrounding a quotation; and (c) we evaluate our models on a publicly available dataset.

\section{Dataset: PDNC}
\label{sec-pdnc}
We briefly describe here the Project Dialogism Novel Corpus \cite{vishnubhotla-etal-2022-project}. 
PDNC consists of 22 full-length English novels, published in the 19th and 20th centuries, annotated with the following information:

\textbf{Characters:} A list of characters in the novel. This includes characters who speak, are addressed to, or referred to multiple times in the novel. Each character is identified by a main name (e.g., \textit{Elizabeth Bennet}), as well as a set of aliases (\textit{Liz, Lizzie, Eliza}). We do not distinguish between the two, and treat each character entity as identifiable by a set of names (so that \textit{Elizabeth Bennet, Liz, Lizzie, Eliza} forms one character entity).
 
 \textbf{Quotations:} Each uttered quotation in the novel is annotated with its speaker and addressee(s); with the referring expression, if any, that indicates who the speaker is; and with internal mentions, \textit{i.e.,} named or pronominal phrases within the quotation that refer to one or more character entities.       
The annotations in PDNC make it ideal for evaluating several aspects of quotation attribution in novels, including named entity recognition, coreference resolution, and speaker attribution.

% \section{Methodology}

\section{Modularization of the Task}
\label{sec-meth}

\textbf{Character identification:} The goal of this sub-task is to build a list of the unique character entities in a novel. Although NER models perform quite well at identifying spans of text that constitute a named entity (here, a character name), the task is complicated by the fact that characters can have multiple aliases in the text. Moreover, some characters may be introduced and referred to only by social titles (\textit{the policeman, the Grand Inquisitor, the little old man, the bystander}). 

\textbf{Coreference resolution:} The goals here are to identify text spans that refer to a character entity (which we refer to as \textit{mentions}) and to link each mention to the correct character entity or entities to which it refers. In addition to mentions that are personal pronouns such as \textit{he, she,} and \textit{them}, literary texts have an abundance of pronominal phrases that reflect relationships between characters, such as \textit{her husband} and \textit{their father}. Such phrases can also occur within quotations uttered by a character (e.g., \textit{my father}), requiring quotation attribution as a prerequisite for complete coreference resolution.

\textbf{Quotation identification:} Perhaps the most straightforward of our sub-tasks, here we identify all text spans in a novel that constitute dialogue, i.e., are uttered by a character entity or entities.

\textbf{Speaker attribution:} Finally, this sub-task links each identified quotation to a named character identity. While most models are designed to solve the more tractable and practical problem of linking quotations to the nearest relevant speaker mention, we subsume the mention--entity linking tasks under the coreference resolution module, equating the two tasks.

\section{Models and Evaluation Metrics}
\label{sec-mod}

We evaluate each of the modules of section \ref{sec-meth} separately. In order not to confound the evaluation with cascading errors, at each step, we ``correct'' the outputs of the automated system from the previous step by using annotations from PDNC. 
% The process is described in more detail in each of the following subsections.

\subsection{Character Identification}
We evaluate two pipelines --- GutenTag and BookNLP --- on their ability to identify the set of characters in a novel, and potentially, the set of aliases for each character. In addition, we also test the NER system from the spaCy\footnote{https://explosion.ai/blog/spacy-v3} module as a proxy for the state-of-the-art in NER that is not trained explicitly for the literary domain.

\textbf{Character recognition (CR):} 
% This metric looks at how well NER systems identify character named entities as such. 
For each novel, we compute the proportion of annotated character entities that are identified as named entities of the category `{\sc person}' \cite{doddington2004automatic}. We use a simple string-matching approach, where we try for either a direct match, or a unique match when common prefixes such as \textit{Mr.\@} and \textit{Sir} are removed. Thus, if a particular novel has $N$ character entities annotated, the NER model outputs a list of $K$ named `{\sc person}' entities, and $K'$ of these entities are in turn matched with $M$ out of the $N$ characters, the CR metric is calculated as $M/N$. 
%     % How many of the unique characters annotated for each novel in PDNC are identified by these systems? \dr{(disregard alias clustering)}
    
\textbf{Character clustering:} We use the clustering evaluation metrics of \textit{homogeneity} (C.Hom), \textit{completeness} (C.Comp), and their harmonic mean, \textit{v-score} to evaluate named entity clusters. Homogeneity (between 0 and 1) is the fraction of named clusters that link to the same character entity; completeness is the number of homogeneous clusters a single entity is distributed over (ideal value of 1).

As an example, consider the case where we have three annotated characters for a novel: \textit{Elizabeth Bennet}, \textit{Mary Bennet}, and \textit{The Queen}. The set of annotated aliases for the characters are \textit{\{Elizabeth Bennet, Eliza, Lizzie, Liz\}}, \textit{\{Mary Bennet, Mary\}}, and \textit{\{The Queen\}}. Say model $M_{1}$ outputs the following entity clusters: \textit{\{Elizabeth Bennet, Eliza\}}, \textit{\{Liz, Lizzie\}} and \textit{\{Mary Bennet, Mary\}}; model $M_{2}$ outputs \textit{\{Elizabeth Bennet, Mary Bennet, Eliza, Mary\}}, \textit{\{Liz, Lizzie\}}. 
Each model has recognized two out of the three characters in our list; this evaluates to a CR score of $2/3$. Each of the three clusters from model $M_{1}$ refers solely to one character entity, resulting in a \textit{homogeneity} score of 1.0. However, these three clusters are formed for only two unique character entities, resulting in a \textit{completeness} score of $1.5$ (\textit{v-score} 0.6). Analogously, model $M_{2}$ has a homogeneity score of 0.5 and a completeness score of 1.0 (\textit{v-score} 0.5).  

\subsection{Coreference Resolution}
We consider two pipelines for coreference resolution: BookNLP (based on \newcite{ju-etal-2018-neural}) and spaCy (based on \newcite{dobrovolskii-2021-word}). 
Given a text, these neural coreference resolution models output a set of clusters, each comprising a set of coreferent mention spans from the input.

Evaluating this module requires annotations that link each mention span in a novel to the character entity referred to. PDNC, unfortunately, contains these mention annotations only for text spans \textit{within} quotations. 

We therefore evaluate coreference resolution only on a subset of the mention spans in a novel, extracted as follows: 
We first identify the set of mention clusters from our models that can be resolved to an annotated character entity, using the character lists from PDNC and the string-matching approach described above. We then prune this to only include those mention spans that are annotated in the PDNC dataset, i.e, mention spans that occur within quotations, and evaluate the accuracy of the resolution.

\textbf{Mention clustering (M-Clus):} We compute the fraction of mention clusters that can be matched to a \textit{unique} (Uniq) annotated character entity rather than to multiple (Mult) or no (None) entities.

\textbf{Mention resolution (M-Res):} For those mention spans within PDNC that are identified by the model and are assigned to a cluster that can be uniquely matched to a character entity (\# Eval), we compute the accuracy of the linking (Acc.).
% \end{itemize}

\subsection{Quotation Identification}
% In line with most prior work, we take the simplest approach to defining \textit{quotations} in a novel: any text span that occurs within quotation marks.
Most models, rule-based or neural, can identify quotation marks and thus quotations. We evaluate how many of such quoted text instances actually constitute \textit{dialogue}, in that they are uttered by one or more characters. Our gold standard is the set of quotations that have been annotated in PDNC, which includes quotations uttered by multiple characters and by unnamed characters such as \textit{a crowd}.

\subsection{Speaker Attribution}

The speaker-attribution part of BookNLP's pipeline is a BERT-based model that uses contextual and positional information to score the BERT embedding for the quotation span against the embeddings of mention spans that occur within a 50-word context window around the quotation; the highest-scoring mention is selected as the speaker. We supplement this approach by limiting the set of candidates to resolved mention spans from the coreference resolution step, thereby directly performing quotation-to-entity linking. As we see from our results, this method, which we refer to as BookNLP+, greatly improves the performance of the speaker attribution model by eliminating spurious candidate spans.

We also evaluate a \textit{sequential prediction model} that predicts the speaker of a quotation simply by looking at the sequence of speakers and mentions that occur in some window around the quotation. We implement this as a one-layer RNN that is fed a sequence of tokens representing the five characters mentioned most recently prior to the quotation text, one character mention that occurs right after, and, optionally, the set of characters mentioned within the quotation. 

\section{Experimental Setup}

We evaluate the models for character identification, coreference resolution, and quotation identification on the entire set of 22 novels in PDNC, since we are neither training nor fine-tuning these on this dataset. For the speaker attribution models, we define the training setup below.

We curate the set of mention candidates for each novel in the following manner: the mention clusters generated by BookNLP are used to extract the set of mention spans that could be successfully resolved to a character entity from the annotated PDNC character lists for each novel. We append to this set the annotated mention spans (within quotations) from PDNC, as well as explicit mention spans --- that is, text spans that directly match a named alias from the character list. 

Overlaps between the three sets are resolved with a priority ranking, whereby PDNC annotations are considered to be more accurate than explicit name matches, which in turn take precedence over the automated coreference resolution model.

We test with 5-fold cross-validation in two ways: splitting the annotated quotations in each novel 80/20 and splitting the set of entire novels 80/20.

\section{Results}

\begin{table}[t]
\adjustbox{width=0.45\textwidth}{
    \begin{tabular}{l rrrrr}
       \textbf{Model}  & \textbf{CR}  & \textbf{C.Hom} & \textbf{C.Comp} & \textbf{v-score} \\
       \hline
       spaCy & 0.81 & 0.16 & 1.02 & 0.27 \\
        GutenTag & 0.60 & 0.98 & 1.33 & 1.12 \\
        BookNLP & 0.85 & 0.86 & 1.18 & 0.99 \\
    \end{tabular}}
    \caption{Character identification: Average scores across all the novels in the dataset. Column headings are defined in the text.  Scores for each individual novel are reported in Appendix \ref{eval_det}.}
 
    \label{tab:char-iden}
\end{table}

% \subsection{Character Identification}
From Table \ref{tab:char-iden}, we see that the neural NER models of spaCy and BookNLP are better at recognizing character names than GutenTag's heuristic system (0.81 and 0.85 vs 0.60).  
However, the strengths of GutenTag's simpler Brown-clustering--based NER system are evident when looking at the homogeneity; when two named entities are assigned as aliases of each other, it is almost always correct. This shows the advantage of document-level named entity clustering as opposed to local span-level mention clustering for character entity recognition. The cluster quality metric, on the other hand, tells us that GutenTag still tends to be conservative with its clustering compared to BookNLP, which nonetheless is a good strategy for the literary domain, where characters often share surnames.

\begin{table}[t]
    \centering
    \adjustbox{width=0.45\textwidth}{
    \begin{tabular}{lrrrr | rr}
        & \multicolumn{4}{c|}{\textbf{M-Clus}} & \multicolumn{2}{c}{\textbf{M-Res}} \\ \hline
        \textbf{Model} & \textbf{\# Clus} & \textbf{Uniq} & \textbf{Mult} & \textbf{None}  & \textbf{\# Eval} & \textbf{Acc.}\\
        \hline
        spaCy & 1503.1 & 0.093 & 0.061 & 0.846 & 499.0 & 0.746\\ 
        %\hline
        BookNLP  & 1662.8 & 0.043 & 0.003 & 0.953 &  1126.6 & 0.774\\
        %\hline
    \end{tabular}}
    \caption{Coreference resolution: All scores are averaged over the 22 novels in PDNC. Column headings are defined in the text.}
   
    \label{tab:coref_eval}
\end{table}

\begin{table}[t]
    \adjustbox{width=0.45\textwidth}{
    \begin{tabular}{l rrr}
    \textbf{Model} & \textbf{Quotations} & \textbf{Novels}\\
    \hline
    BookNLP-OG & 0.40 & 0.40 \\
   % \hline
      BookNLP+ (LitBank) & 0.62 & 0.61 \\
      Seq-RNN & 0.72 & 0.64\\
      BookNLP+ (PDNC) & 0.78 & 0.68 \\
      %\hline
    \end{tabular}}
    \caption{Accuracy on speaker attribution for the end-to-end BookNLP model (BookNLP-OG), the restricted model with only resolved mention spans as candidates (row 2), the sequential prediction model, and the restricted model trained on PDNC, for the Quotations and the entire Novels cross-validation split.}
    \label{tab:spk-att}
\end{table}

% \subsection{Coreference Resolution}
Performance of these models on the coreference resolution task is significantly lower (Table~\ref{tab:coref_eval}). A majority of the mention clusters from both BookNLP and spaCy's coreference resolution modules end up as unresolved clusters, with no containing named identifier that could be linked to a PDNC character entity. However, when we evaluate mention-to-entity linking on the subset of clusters that \textit{can} be resolved, both systems achieve accuracy scores of close to 0.78, although spaCy is able to resolve far fewer mentions (499 vs 1127).

The importance of the character identification and coreference resolution tasks can be quantified by looking the performance of the speaker attribution models (Table~\ref{tab:spk-att}).
% (accuracy of quotation identification is 0.98 with a simple quotation mark identifier) 
The end-to-end pretrained BookNLP pipeline, when evaluated on the set of PDNC quotations (which were identified with accuracy of 0.94), achieves an accuracy of 0.42. When we restrict the set of candidate mentions for each quotation to only those spans that can be resolved to a unique character entity, the attribution accuracy increases to 0.61. 
% \dr{That's not in the table.} 
However, the RNN model still beats this performance with an accuracy of 0.72 on the random data split. When BookNLP's contextual model is trained on data from PDNC, its accuracy improves to 0.78. These scores drop to 0.63 and 0.68 for the entire-novel split, where we have the disadvantage of being restricted only to patterns of mention sequences, and not speakers.

\section{Analysis}
We  briefly go over some qualitative analyses of the errors made by models in the different sub-tasks, which serves to highlight the challenges presented by literary text and opportunities for future research.

\paragraph{Character Identification and Coreference Resolution: }We manually examine the mention clusters identified by our coreference resolution modules that could not be matched a unique character entity as annotated in PDNC.

We find that, by far, the most common error is conflating characters with the same surname or family name within a novel. For example, several of the women characters in these novels are often referred to by the names of their husbands or fathers, prefixed with a honorific such as {\it Mrs.\@} or {\it Miss}. Thus \textit{Mrs.\@ Archer} refers to \textit{May Welland} in \textit{The Age of Innocence} and \textit{Miss Woodhouse} refers to \textit{Emma Woodhouse} in \textit{Emma}. However, a surname without a title, such as \textit{Archer} or \textit{Woodhouse}, generally refers to the corresponding male character. This results in the formation of mention clusters that take the spans \textit{Miss Woodhouse} and \textit{Woodhouse} to be coreferent, despite being different character entities. We see similar issues with father--son character pairs, such as \textit{George Emerson} and \textit{Mr.\@ Emerson} in \textit{A Room With A View}, and with character pairs that are siblings.

\begin{table}[t]
    \centering
    \adjustbox{width=0.45\textwidth}{
    \begin{tabular}{lrr | rr}
        & \multicolumn{2}{c|}{\textbf{Quotations}} & \multicolumn{2}{c}{\textbf{Novels}} \\ \hline
        \textbf{Model} & \textbf{Exp.} & \textbf{Rest} & \textbf{Exp.}  & \textbf{Rest}\\
        \hline
        BookNLP-OG & 0.64 & 0.28 & 0.63 & 0.28\\ 
        %\hline
        BookNLP+ (LitBank)  & 0.93 & 0.47 & 0.95 & 0.43 \\
        Seq-RNN  & 0.85 & 0.65 & 0.76 & 0.57 \\ 
        BookNLP+ (PDNC)  & 0.98 & 0.70 & 0.97 & 0.53 \\
        %\hline
    \end{tabular}}
    \caption{Attribution accuracy for the speaker attribution models, broken down by quotation type, for the Quotations and Novels cross-validation splits. Column Exp. refers to explicit quotations, and column Rest refers to implicit and anaphoric quotations.}

    \label{tab:qtype-perf}
\end{table}

\paragraph{Speaker Attribution: }
We first quantify the proportion of quotations attributed to a mention cluster that cannot be resolved to a named character entity with the end-to-end application of the BookNLP pipeline.

On average, 47.7\% of identified quotations are assigned to an unresolved mention cluster as the speaker. The range of this value varies from as low as 12.5\% (\textit{The Invisible Man}) to as high as 78.7\% (\textit{Northanger Abbey}).
A majority of these unresolved attributions occur with implicit and anaphoric quotations (76.2\%),  where the speaker is not explicitly indicated by a referring expression such as \textit{Elizabeth said}, as opposed to explicit quotations (23.8\%).
% How many of these are anaphoric, as opposed to implicit quotations?

% What is the performance of the final model on explicit, anaphoric, and implicit quotations?
In Table \ref{tab:qtype-perf}, we break down the performance of the speaker attribution models by quotation type. We see that even our local context--based RNN model is able to identify the speaker of explicit quotations with a relatively high accuracy, and that the speaker for non-explicit quotations can also generally be modeled using the sequence of 5--6 characters mentioned in the vicinity of the quotation. The transformer-based models are of course able to use this local context more effectively by making use of linguistic cues and non-linear patterns of mentions and speakers in the surrounding text. Still, our best performing model achieves an accuracy of only 0.53 on implicit and anaphoric quotations when applied to novels unseen in the training set (the Novels split).
%\todo{(TBD, IPR)}

\section{Conclusions and Future Work}
In this work, we quantitatively evaluated the key components of a functional quotation attribution system. We showed that the initial task of recognizing characters and their aliases in a novel remains quite a challenge, but doing so greatly improves the performance of speaker attribution by limiting the set of candidate speakers. However, with existing coreference resolution systems, a large portion of mention clusters (around 90\%) remain unresolved, so this remains a problem for new research.

% \section{Conclusions and Future Work}

\section*{Limitations}
There is much variation in literary writing and narrative styles, and our work here deals with a small, curated subset of this domain. The novels we analyze are all in the English language, and were published between the early 19th and early 20th centuries. The authors and novels themselves are drawn from what is considered to be the established literary canon, and are not necessarily representative of all the works of that era, let alone literary works of other eras. 
The texts we analyze are largely uniform in narrative style. We limit ourselves to only those quotations that are explicitly indicated as such in the text by quotation marks, thereby eliminating more-complex styles such as free indirect discourse \cite{Brooke2016UsingMO} and stream-of-consciousness novels. We do not deal with nuances such as letters and diary entries nor quotations within quotations. The models we analyze for named entity recognition and coreference resolution use a fixed, binary formulation of the gender information conveyed by pronominal terms. Though the development of fairer, more representative models is constrained by current datasets, we note that there is encouraging progress being made in this area \cite{bamman2020annotated, yoder-etal-2021-fanfictionnlp}. 
% ACL 2023 requires all submissions to have a section titled ``Limitations'', for discussing the limitations of the paper as a complement to the discussion of strengths in the main text. This section should occur after the conclusion, but before the references. It will not count towards the page limit.
% The discussion of limitations is mandatory. Papers without a limitation section will be desk-rejected without review.

% While we are open to different types of limitations, just mentioning that a set of results have been shown for English only probably does not reflect what we expect. 
% Mentioning that the method works mostly for languages with limited morphology, like English, is a much better alternative.
% In addition, limitations such as low scalability to long text, the requirement of large GPU resources, or other things that inspire crucial further investigation are welcome.

% \section*{Ethics Statement}
% Scientific work published at ACL 2023 must comply with the ACL Ethics Policy.\footnote{\url{https://www.aclweb.org/portal/content/acl-code-ethics}} We encourage all authors to include an explicit ethics statement on the broader impact of the work, or other ethical considerations after the conclusion but before the references. The ethics statement will not count toward the page limit (8 pages for long, 4 pages for short papers).

% \section*{Acknowledgements}

\bibliography{anthology,custom}
\bibliographystyle{acl_natbib}

\appendix

\section{Implementation Details}

The BookNLP pipeline is available to use as a Python package, as is spaCy, with pretrained models for coreference resolution and speaker attribution. For the former, these models are trained on the LitBank corpus, which is a dataset from the literary domain. We use these pretrained models to evaluate performance on the character identification and coreference resolution tasks. GutenTag can be run either via a Web interface or a command-line executable (requiring Python~2). It was designed to interface with texts from the Project Gutenberg corpus. Some of the novels in PDNC were not found in GutenTag's predefined database of texts, so we exclude these when reporting average performance metrics. 
\section{Results by Novel}
\label{eval_det}
Tables \ref{tab:char_rec} and \ref{tab:spacy-chars} show for each novel in PDNC the per-model results for character identification that are summarized in Table \ref{tab:char-iden}.

\begin{table*}
    \centering
    \adjustbox{width=\textwidth}{%
    \begin{tabular}{lr|rrrrr|rrrrr}
    & \multicolumn{5}{c}{\textbf{BookNLP}} & \multicolumn{5}{c}{\textbf{GutenTag}}\\ 
    \textbf{Novel}&\textbf{\# Chars}& \textbf{CR}&\textbf{\# Clus} & \textbf{C.Hom}&\textbf{C.Comp} & \textbf{v-score} & \textbf{CR}&\textbf{\# Clus} & \textbf{C.Hom}&\textbf{C.Comp} & \textbf{v-score}\\
    \hline
    \textit{A Room With A View}&63&0.83&60&0.95&1.19&1.06&0.48&35&1.00&1.17&1.08\\
\textit{The Age of Innocence} &55&0.84&48&0.81&1.26&0.99&0.64&49&1.00&1.40&1.17\\
\textit{Alice's Adventures in Wonderland} &51&0.67&34&0.97&1.03&1.00&0.25&14&1.00&1.08&1.04\\
\textit{Anne of Green Gables}&113&0.87&102&0.92&1.08&0.99&0.19&25&1.00&1.14&1.06\\
\textit{Daisy Miller}&10&1.00&13&1.00&1.30&1.13&0.80&12&1.00&1.50&1.20\\
\textit{Emma}&18&0.89&17&0.71&1.09&0.86&0.89&27&1.00&1.69&1.26\\
\textit{A Handful of Dust}&104&0.82&94&0.89&1.15&1.01&$-$&$-$&$-$&$-$&$-$\\
\textit{Howards End}&55&0.95&64&0.89&1.27&1.05&0.49&33&0.97&1.23&1.08\\
\textit{Night and Day}&50&0.94&53&0.77&1.17&0.93&0.62&40&0.97&1.30&1.11\\
\textit{Northanger Abbey}&20&0.90&12&0.75&1.00&0.86&0.85&23&0.96&1.29&1.10\\
\textit{Persuasion}&35&0.86&29&0.79&1.28&0.98&0.77&28&0.96&1.08&1.02\\
\textit{Pride and Prejudice}&74&0.81&62&0.85&1.10&0.96&0.35&30&0.90&1.35&1.08\\
\textit{Sense and Sensibility}&24&0.83&25&0.56&1.17&0.76&0.79&26&0.96&1.39&1.14\\
\textit{The Sign of the Four}&35&0.94&32&0.72&1.05&0.85&0.60&28&1.00&1.33&1.14\\
\textit{The Awakening}&22&0.82&17&0.88&1.07&0.97&0.77&21&0.95&1.25&1.08\\
\textit{The Gambler} &27&0.70&22&0.91&1.18&1.03&0.59&22&1.00&1.38&1.16\\
\textit{The Invisible Man}&31&0.94&40&0.95&1.36&1.12&0.61&32&1.00&1.68&1.25\\
\textit{The Man Who Was Thursday}&30&0.80&35&0.97&1.55&1.19&0.53&23&1.00&1.44&1.18\\
\textit{The Mysterious Affair at Styles} &30&0.80&25&0.88&1.05&0.96&0.70&28&0.96&1.35&1.12\\
\textit{The Picture of Dorian Gray} &43&0.88&43&0.98&1.14&1.05&0.56&27&1.00&1.12&1.06\\
\textit{The Sport of the Gods} &37&0.81&34&0.94&1.23&1.07&0.54&28&0.96&1.50&1.17\\
\textit{The Sun Also Rises} &51&0.86&51&0.96&1.23&1.08&$-$&$-$&$-$&$-$&$-$\\
\hline
\textbf{Mean} & \textbf{44.5} & \textbf{0.85} & \textbf{41.45} & \textbf{0.86} & \textbf{1.18} & \textbf{0.99} & \textbf{0.60} & \textbf{27.55} & \textbf{0.98} & \textbf{1.33} & \textbf{1.12} 
\\

    \hline

    \end{tabular}}
    \caption{Results of character identification for each novel with BookNLP and GutenTag. `\# Chars' is the number of characters in the novel.  Other headers are the same as in Table \ref{tab:char-iden}.}
    \label{tab:char_rec}
\end{table*}

\begin{table*}[]
    \centering {\small
    \begin{tabular}{lr|rrrrr}
    \textbf{Novel}&\textbf{\# Chars}&\textbf{CR} & \textbf{\# Clus} & \textbf{C.Hom} &\textbf{C.Comp}&\textbf{v-score}\\
    \hline
    \textit{A Room With A View}&63&0.78&64&0.33&1.24&0.52\\
    \textit{The Age of Innocence}&55&0.85&90&0.04&1.00&0.09\\
    \textit{Alice's Adventures in Wonderland}&51&0.80&44&0.39&1.00&0.56\\
    \textit{Anne of Green Gables}&113&0.69&98&0.24&1.04&0.40\\
    \textit{Daisy Miller}&10&0.90&3&0.00&0.00&0.00\\
    \textit{Emma}&18&0.89&14&0.07&1.00&0.13\\
    \textit{A Handful of Dust}&104&0.71&85&0.26&1.00&0.41\\
    \textit{Howards End}&55&0.84&72&0.18&1.08&0.31\\
    \textit{Night and Day}&50&0.88&52&0.15&1.00&0.27\\
    \textit{Northanger Abbey}&20&0.90&15&0.07&1.00&0.12\\
    \textit{Persuasion}&35&0.89&36&0.06&1.00&0.11\\
    \textit{Pride and Prejudice}&74&0.68&78&0.17&1.00&0.29\\
    \textit{Sense and Sensibility}&24&0.83&21&0.10&1.00&0.17\\
    \textit{The Sign of the Four}&35&0.80&40&0.05&1.00&0.10\\
    \textit{The Awakening}&22&0.86&24&0.12&1.00&0.22\\
    \textit{The Gambler}&27&0.74&18&0.22&1.00&0.36\\
    \textit{The Invisible Man}&31&0.84&37&0.22&1.00&0.36\\
    \textit{The Man Who Was Thursday}&30&0.73&26&0.19&1.00&0.32\\
    \textit{The Mysterious Affair at Styles}&30&0.87&29&0.10&1.00&0.19\\
    \textit{The Picture of Dorian Gray}&43&0.86&32&0.19&1.00&0.32\\
    \textit{The Sport of the Gods}&37&0.81&43&0.12&1.00&0.21\\
    \textit{The Sun Also Rises}&51&0.82&56&0.32&1.12&0.50\\
    \hline
\textbf{Mean} & \textbf{44.5} & \textbf{0.81} & \textbf{44.40} & \textbf{0.16} & \textbf{1.02} & \textbf{0.27}
    \end{tabular}
}
    \caption{Results of character identification for each novel with spaCy. `\# Chars' is the number of characters in the novel.  Other headers are the same as in Table \ref{tab:char-iden}.}
    \label{tab:spacy-chars}
\end{table*}

\end{document}